\documentclass[letterpaper, 10 pt, journal, twoside]{IEEEtran}

\usepackage{graphics}

\usepackage{amsfonts}
\usepackage{amsmath}
\DeclareMathOperator*{\argmin}{\arg\!\min}
\usepackage{amssymb}
\usepackage[colorlinks, linkcolor=blue, citecolor=blue]{hyperref}
\usepackage{bm}

\usepackage{algorithm}
\usepackage{url}
\usepackage{algpseudocode}
\usepackage{epsfig}
\usepackage{graphicx}
\usepackage{caption}
\usepackage{subfigure}
\usepackage{mathrsfs}
\usepackage{array}
\usepackage{multirow}
\usepackage{multicol}
\usepackage{booktabs}
\usepackage{makecell}
\usepackage{xcolor}
\usepackage{amsfonts}
\usepackage{gensymb}
\usepackage{biblatex}
\usepackage{flushend}
\usepackage[flushleft]{threeparttable}

\title{Kinematic Motion Retargeting via Neural Latent Optimization for Learning Sign Language}

\author{Haodong Zhang, Weijie Li, Jiangpin Liu, Zexi Chen, Yuxiang Cui, Yue Wang, Rong Xiong
\thanks{All authors are with the State Key Laboratory of Industrial Control and Technology, Zhejiang University, Hangzhou, P.R. China. Yue Wang and Rong Xiong are the corresponding authors. {\tt\small wangyue@iipc.zju.edu.cn, rxiong@zju.edu.cn}}
}

\begin{document}

\maketitle
\thispagestyle{empty}
\pagestyle{empty}

\begin{abstract}

Motion retargeting from a human demonstration to a robot is an effective way to reduce the professional requirements and workload of robot programming, but faces the challenges resulting from the differences between humans and robots. Traditional optimization-based methods are time-consuming and rely heavily on good initialization, while recent studies using feedforward neural networks suffer from poor generalization to unseen motions. Moreover, they neglect the topological information in human skeletons and robot structures. In this paper, we propose a novel neural latent optimization approach to address these problems. Latent optimization utilizes a decoder to establish a mapping between the latent space and the robot motion space. Afterward, the retargeting results that satisfy robot constraints can be obtained by searching for the optimal latent vector. Alongside with latent optimization, neural initialization exploits an encoder to provide a better initialization for faster and better convergence of optimization. Both the human skeleton and the robot structure are modeled as graphs to make better use of topological information. We perform experiments on retargeting Chinese sign language, which involves two arms and two hands, with additional requirements on the relative relationships among joints. Experiments include retargeting various human demonstrations to YuMi, NAO, and Pepper in the simulation environment and to YuMi in the real-world environment. Both efficiency and accuracy of the proposed method are verified.

\end{abstract}



\section{INTRODUCTION}

Motion retargeting simplifies robot programming by learning human demonstrations, which can effectively reduce the requirement of programming expertise and enable rapid learning of complex robot movements. In this paper, we focus on generating kinematically feasible robot motions, which can help robots express specific information or emotion with body language. Nowadays, it has been applied to humanoid robots in entertainment parks \cite{hoshyari2019vibration} and sign language robots for communication with the hearing impaired \cite{liang2021dynamic}. Moreover, it could be used for service robots in museums or restaurants to interact with people using body movements. {Specifically, we perform motion retargeting on the task of unseen sign language, which includes complex dual-arm movements and finger movements. However, it still remains an ongoing challenge due to the differences between humans and robots. Even for humanoid manipulators with structures similar to human beings, there exist differences in degrees of freedom, kinematic parameters and physical constraints.} Together with various requirements of similarity, safety and rapidity, the problem becomes difficult to solve.

Previous work has been developed to address this problem. Direct mapping \cite{penco2018robust} transforms human motions through a human-defined mapping relationship, but is hard to define manually. Methods based on inverse kinematics \cite{asfour2003human} have been used to keep the end effector positions of the robot consistent with those of the human, whereas it does not consider the robot constraints and the similarity of other joints. To overcome these problems, optimization-based methods have been proposed to find the optimal solution that maximizes the motion similarity and satisfies the robot execution capability \cite{liang2021dynamic,choi2019towards,wang2017generative}. It is usually achieved by defining and optimizing an objective function with constraints. Although these methods are able to produce promising retargeting results, they have to take extensive time to optimize each motion and poor initialization may lead to a bad local minimum.

\begin{figure}[t]
\centering
\vspace{6pt}
\includegraphics[width=\linewidth]{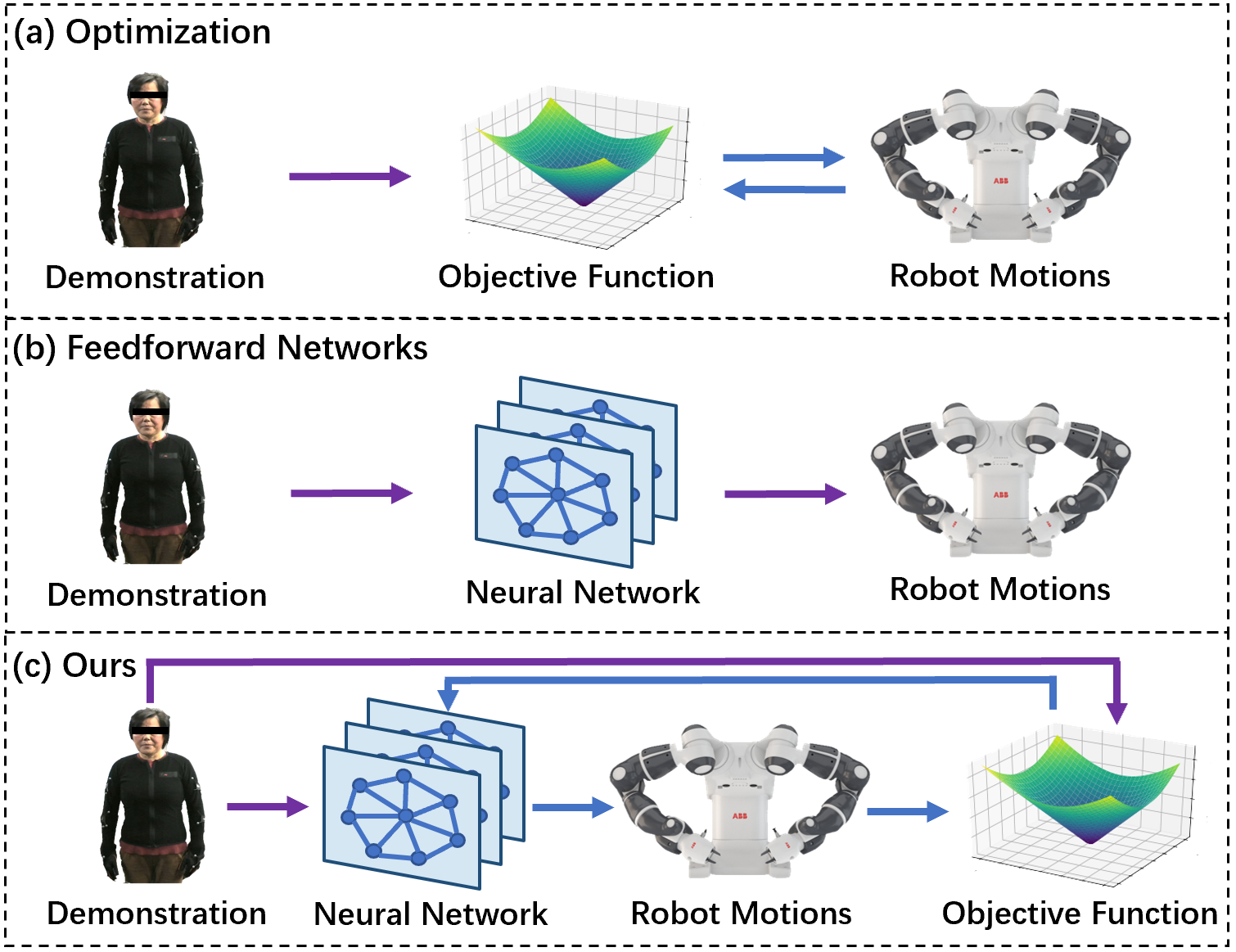}
\caption{Illustration of different motion retargeting methods. (a) Optimization methods. (b) Feedforward network methods. (c) Our method. Purple arrows denote feedforward pass, and blue arrows denote iterative pass.}
\label{introduction}
\vspace{-18pt}
\end{figure}

Latent motion is one of the research hotspots in recent years, and there is some work applying latent space to different tasks \cite{ichter2019robot,watter2015embed,merel2018neural}. Ichter et al. \cite{ichter2019robot} introduced an RRT-based algorithm to plan motions directly in the latent space for visual planning and humanoid robot planning. Watter et al. \cite{watter2015embed} proposed a locally linear latent dynamics model for control from raw pixel images. However, these methods cannot be directly applied in the task of motion retargeting, since they do not take into account motion similarity. A reinforcement learning based method \cite{kim2020c} tends to address the problem by designing rewards for retargeted motions. Choi et al. \cite{choi2020nonparametric} proposed a data-driven motion retargeting method that utilizes the results of an optimization-based method as ground truth for training. However, due to limited training data, these methods may perform poorly on unseen motions, which results in inaccurate or infeasible robot movements.

In this paper, we propose a neural latent optimization method that leverages the advantages of both neural networks and optimization. Specifically, we first formulate motion retargeting as a constrained optimization problem and transform it to an unconstrained one with the help of a deep decoder. The decoder learns the mapping from the latent space to the robot motion space. For any unseen motion, we further search for the optimal latent code that minimizes the objective function. To accelerate the optimization process and help converge to better results, we exploit a deep encoder to generate better initial values for the latent code. The human skeleton and robot structure are modeled as graphs, which can make better use of topological information and have better generalization. The graph encoder and graph decoder are trained end-to-end. To the best of our knowledge, this is the first work utilizing neural latent optimization for motion retargeting from a human to a robot.

Our contributions can be summarized as follows:

\begin{itemize}
	

    \item {propose a novel kinematic motion retargeting framework by integrating latent optimization and neural initialization, and build a Chinese sign language dataset for human-robot motion retargeting.}

	

    \item {introduce latent optimization to improve the performance on unseen motions, and design neural initialization to provide better initial values for the latent vector, which helps converge faster and better.}
	
	\item model the human skeleton and robot structure as graphs, which can capture the inherent information of the topological structure and have better generalization.
	

	
\end{itemize}


\section{Related Work}

\subsection{Learning from Demonstration}

Learning from demonstration has been a fundamental problem in robotics research and holds great promise for industrial applications, since it enables robots to mimic human skills without programming \cite{zhu2018robot}. Recent studies have shown progress in learning from demonstration. Hamaya et al. \cite{hamya} proposed a method for learning soft robotic assembly strategies by designing reward functions from human demonstrations. Lee et al. \cite{leemix} proposed a mixed generative adversarial imitation learning approach, which incorporates both expert demonstrations and negative demonstrations. Cai et al. \cite{cai2020inferring} introduced a framework to learn skills from human demonstrations in the form of geometric nullspaces, which infers a parameterized constraint-based skill model independent of the robot kinematics or environment. Different from these approaches, we aim to address the human-robot motion retargeting problem, where the robot mimics human motions while maintaining motion similarity and satisfying the robot's kinematic constraints.


\subsection{Optimization-based Motion Retargeting}

Previous work has attempted to solve the kinematic motion retargeting problem with optimization. By optimizing a predefined objective function of kinematics similarity in an iterative manner, these approaches minimize the gap between robot motions and human demonstrations. Liang et al. \cite{liang2021dynamic} proposed a motion retargeting method that leverages graph optimization and Dynamic Movement Primitives. Wang et al. \cite{wang2017generative} presented a generative framework for motion retargeting using a single depth sensor. Choi et al. \cite{choi2019towards} designed a motion retargeting pipeline composing of motion retarget descriptions, optimization, inverse kinematics and trajectories post-processing. However, these methods are computationally expensive and require a long iterative process for optimization. Moreover, due to the complex objective function, they are also prone to fall into a bad local minimum without a good initial value. As a comparison, we propose to optimize on the latent space rather than the robot motion space by utilizing a graph decoder. We also introduce a graph encoder to provide a better initial value for optimization, which can accelerate the iterative process.

\subsection{Learning-based Motion Retargeting}

Most learning-based methods focus on motion retargeting of animation characters in computer graphics. Aberman et al. \cite{aberman2020skeleton} introduced differentiable skeleton-aware convolution, pooling and unpooling operators for unpaired motion retargeting. They also presented a method for retargeting video-captured motion \cite{aberman2019learning}. Villegas et al. \cite{villegas2018neural} proposed a recurrent neural network with a forward kinematic layer and cycle consistency based adversarial training objective. Although these methods have achieved impressive results in animation characters, they cannot be directly applied to human-robot motion retargeting because they do not take into account the robot's kinematic constraints and would produce infeasible robot movements. For human-robot kinematic motion retargeting, Kim et al. \cite{kim2020c} developed a cyclic three-phase optimization method based on deep reinforcement learning. Choi et al. \cite{choi2020nonparametric} proposed a data-driven motion retargeting method combining the nonparametric regression and deep latent variable modeling. However, since the quantity and diversity of training data are limited, these methods may not perform well in unseen motions, resulting in inaccurate or infeasible robot motions. As a comparison, our method further optimizes the latent code for any unseen motion, which can improve the performance and combine the advantages of neural networks and optimization. Moreover, our method takes into account not only arm movements, but also complex finger movements.
\section{Problem Statement}



{In this section, we focus on the problem of motion retargeting which requires to generate kinematically feasible robot motions, and formulate the problem as minimizing an objective function, which considers motion similarity and robot kinematic constraints.} Suppose we have a frame of the human demonstration, denoted as $D$, which represents the poses of all human joints at this frame. Then the goal of motion retargeting is to minimize the difference between the robot motion and the human demonstration. While there exist several different ways to represent the robot motion, we propose to use joint angles instead of joint positions, and then calculate joint positions with a differentiable forward kinematics module. Such a representation can avoid changes of joint length and multiple solutions of inverse kinematics. Hence the robot motion can be represented by $K(\boldsymbol{r})$, where ${\boldsymbol{r}}$ is the robot joint angles and $K$ is the forward kinematics module. Finally, the goal of motion retargeting can be formulated as follows:
\begin{equation}
\label{object_func}
\begin{aligned}
& \underset{\boldsymbol{r}}{\text{minimize}}
& & L(D,K(\boldsymbol{r})) \\
& \text{subject to}
& & \boldsymbol{r}_{lower} \leq \boldsymbol{r} \leq \boldsymbol{r}_{upper}
\end{aligned}
\end{equation}

\noindent where $L$ is the objective function that measures the difference between robot motions and human demonstrations, $\boldsymbol{r}_{upper}$ and $\boldsymbol{r}_{lower}$ are the upper and lower limits of the joint angle.

{In order to encourage robot motions to be as similar as possible to human demonstrations and kinematically feasible, the overall objective function is composed of five terms}: end effector loss $L_{ee}$, orientation loss $L_{ori}$, elbow loss $L_{elb}$, finger loss $L_{fin}$ and collision loss $L_{col}$, {where $\lambda_{ee}$, $\lambda_{ori}$, $\lambda_{elb}$, $\lambda_{fin}$ and $\lambda_{col}$ are their weights with sizes of 1000, 100, 100, 100 and 1000, respectively.} The overall objective function needs to be modified to adapt to other tasks. For example, for the task of object grasping and manipulation, the end effector loss and the finger loss should be redesigned and an additional pose estimation module is needed.

\vspace{-10pt}
\begin{equation}
\begin{split}
    L = \lambda_{ee}L_{ee} + \lambda_{ori}L_{ori} + \lambda_{elb}L_{elb}\\ + \lambda_{fin}L_{fin} + \lambda_{col}L_{col}
\end{split}
\end{equation}

\noindent\textbf{End Effector Loss:} The end effector loss $L_{ee}$ encourages the robot to match end effector positions of human demonstrations, and compares the difference of the normalized end effector positions, using mean square error. {The normalization coefficient is the actual length from the shoulder to the wrist.} Let $\boldsymbol{p}_j$ and $l_j$ be the position and the normalization coefficient of the end effector $j$, and $\boldsymbol{\hat{p}}_j$ and $\hat{l}_j$ be the corresponding variables of the demonstration. Then $L_{ee}$ is defined as:

\begin{equation}
    L_{ee} = \sum_j||\frac{\boldsymbol{p}_j}{l_j}-\frac{\boldsymbol{\hat{p}}_j}{\hat{l}_j}||_2^2
\end{equation}

\noindent\textbf{Orientation Loss:} The orientation loss $L_{ori}$ is calculated by comparing the difference of end effector orientations, using mean square loss as well. Let $\boldsymbol{R}_j$ and $\boldsymbol{\hat{R}}_j$ be the end effector rotation matrix of the robot and the human respectively. Then $L_{ori}$ is given as:

\begin{equation}
    L_{ori} = \sum_j||\boldsymbol{R}_j-\boldsymbol{\hat{R}}_j||_2^2
\end{equation}

\noindent\textbf{Elbow Loss:} The elbow loss $L_{elb}$ encourages the movements of other joints to be similar to the demonstration. It is calculated by comparing the difference of the normalized vectors from the elbow to the wrist. {The normalization coefficient is the actual length from the elbow to the wrist.} Let $\boldsymbol{p}_j^{wrist}$, $\boldsymbol{p}_j^{elbow}$, and $l_{we}$ be the wrist position, the elbow position, and the normalization coefficient of the robot's arm $j$, and $\boldsymbol{\hat{p}}_j^{wrist}$, $\boldsymbol{\hat{p}}_j^{elbow}$, and $\hat{l}_{we}$ be the corresponding variables of the demonstration. Then $L_{elb}$ is calculated as:
\begin{equation}
    L_{elb} = \sum_j||\frac{\boldsymbol{p}_j^{wrist}-\boldsymbol{p}_j^{elbow}}{l_{we}}-\frac{\boldsymbol{\hat{p}}_j^{wrist}-\boldsymbol{\hat{p}}_j^{elbow}}{\hat{l}_{we}}||_2^2
\end{equation}

\noindent\textbf{Finger Loss:} The finger loss $L_{fin}$ encourages the robot to match the finger movements of the human, which are an important part of sign language. It compares the vectors from the metacarpophalangeal joint to the fingertip, normalized by the finger length. Let $\boldsymbol{p}_j^{tip}$, $\boldsymbol{p}_j^{meta}$, and $l_{tm}$ be the fingertip position, the metacarpophalangeal joint position, and the normalization coefficient of the robot's finger $j$, and $\boldsymbol{\hat{p}}_j^{tip}$, $\boldsymbol{\hat{p}}_j^{meta}$, and $\hat{l}_{tm}$ be the corresponding variables of the demonstration. Then $L_{fin}$ is defined as:
\begin{equation}
    L_{fin} = \sum_j||\frac{\boldsymbol{p}_j^{tip}-\boldsymbol{p}_j^{meta}}{l_{tm}}-\frac{\boldsymbol{\hat{p}}_j^{tip}-\boldsymbol{\hat{p}}_j^{meta}}{\hat{l}_{tm}}||_2^2
\end{equation}

\noindent\textbf{Collision Loss:} The collision loss $L_{col}$ is designed to penalize robot motions that result in collisions. We model the links of the robotic arms as capsules and calculate the distance between pairs of capsules. If this distance is less than the minimum distance without collision, the loss is calculated. Let $d_{i,j}$ be the distance between the capsule $i$ and the capsule $j$, and $d_{min}$ be the threshold of no collision. Then $L_{col}$ is given as:

\begin{equation}
    L_{col} = \sum_{d_{i,j}<d_{min}}e^{-d_{i,j}^2}
\end{equation}



\begin{figure*}[htbp]
\centering
\includegraphics[width=0.8\linewidth]{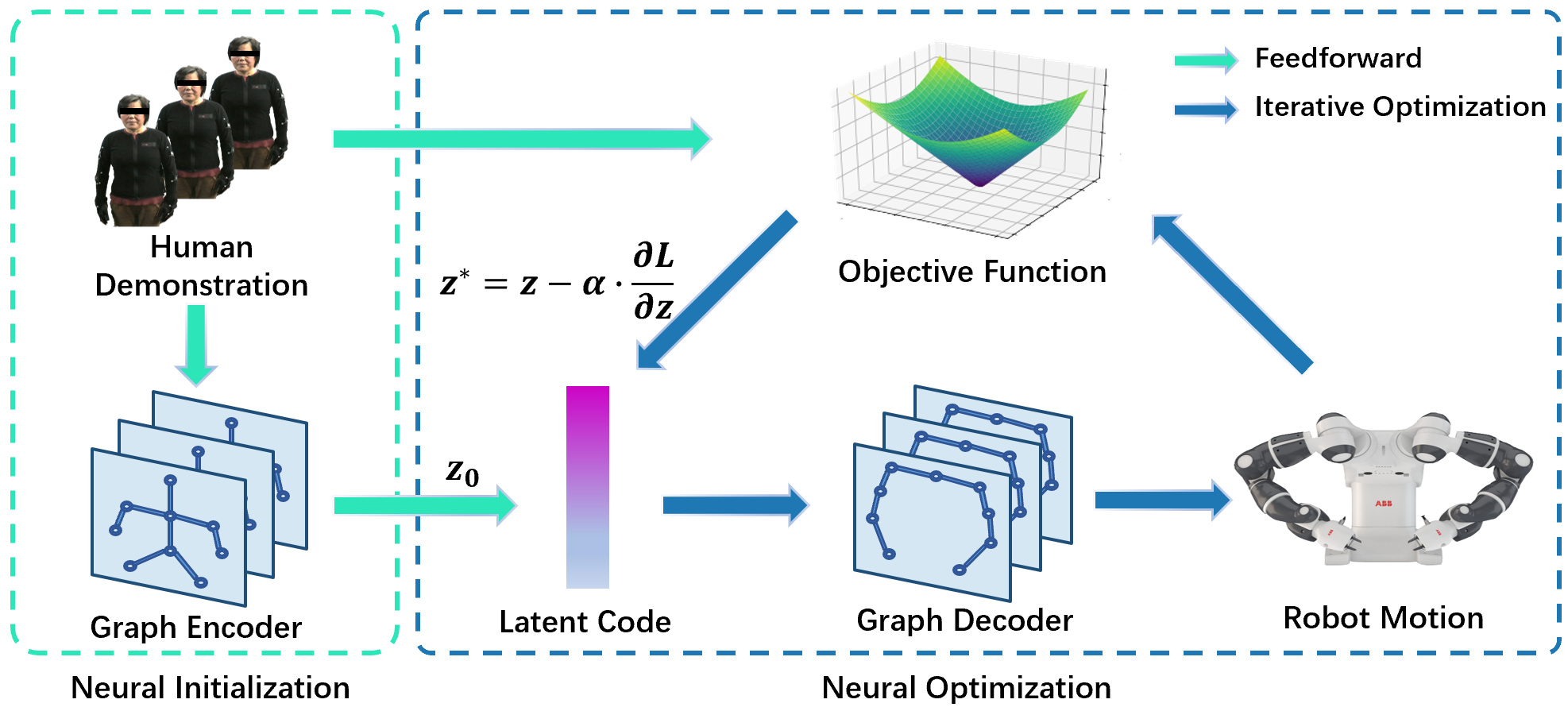}
\caption{Overall framework of our method. We establish a mapping relationship between the latent space and the robot motion space with a graph decoder. Utilizing the difference between human demonstrations and robot motions, we search for the optimal latent code $\boldsymbol{z}^*$ that minimizes the gap by gradient descent. The initial value of the latent code $\boldsymbol{z}_0$ is provided by neural initialization instead of random initialization, which improves the convergence speed.}
\label{framework}
\vspace{-0.4cm}
\end{figure*}

\section{Neural Latent Optimization}

A class of traditional methods attempt to solve the constrained optimization problem in Eq. \ref{object_func} by adding the joint limit loss $L_{lim}$ as a soft constraint. The new objective function $L_{new}$ is formulated as Eq. \ref{new_obj}. However, the objective function is highly nonconvex, making these methods difficult to solve well. The main reasons are as follows: 1) poor initialization may lead to a bad local minimum as well as a time-consuming optimization process; 2) the optimization results may exceed the joint limits because $L_{lim}$ may not be minimized to zero.

\vspace{-5pt}
\begin{equation}
\begin{split}
\label{new_obj}
    L_{new} = L + \lambda_{lim}L_{lim}
\end{split}
\end{equation}

In contrast, our method aims to overcome these drawbacks with the help of latent optimization and neural initialization. As shown in Fig. \ref{framework}, we propose to optimize in the latent space instead of the robot motion space, where the latent vectors are mapped to hard constrained robot motions by a graph decoder. Furthermore, we utilize a deep graph encoder to provide better initialization, which improves the convergence speed and results.

\subsection{Latent Optimization}

We first describe how to convert the constrained optimization problem into an unconstrained one. In order to constrain the output joint angles, a practical way is to use a nonlinear activation function. Here we use $tanh$ to bound the output to a finite range. Then we linearly remap the output to the upper and lower limits of the robot's joint angles, which ensures that the joint limit constraints are satisfied. 
\noindent\textbf{Auto-decoder:} To improve the performance on unseen motions, we propose to employ an auto-decoder \cite{park2019deepsdf} architecture, where we choose to optimize the objective function in the latent space rather than the robot motion space. To achieve this, we introduce a latent vector $\boldsymbol{z}$ as a latent embedding of the robot motion. Different latent codes correspond to different robot motions. A deep decoder $f_{\psi}$ is utilized to establish the connection between the latent space and the robot motion space, where $\psi$ is the learnable parameters of the decoder. The goal of the decoder is to map the latent vector to the corresponding robot motion, which adheres the joint limit constraints with the help of the nonlinear activation function. A simple approach to train the decoder is to jointly optimize the parameters of the decoder $f_{\psi}$ and the latent vector $\boldsymbol{z}$ by gradient descent as Eq. \ref{autodecoder}. The model simultaneously learns the mapping function and finds the optimal latent code for each motion. The regularization term is designed to encourage the latent space to be close to the a multivariate Gaussian distribution with a mean of 0 and a variance of $\sigma$. {The variance $\sigma$ is chosen as 1, similar to the Variational Auto-encoders \cite{kingma2013auto}. More details of the motivation of using latent space can be found in the Appendix \cite{zhang2021kinematic}.}


\vspace{-10pt}
\begin{equation}
\label{autodecoder}
    \min_{\boldsymbol{z},\psi}L(D,K(f_{\psi}(\boldsymbol{z})))+\frac{1}{\sigma^2}||\boldsymbol{z}||_2^2\\
\end{equation}






\noindent\textbf{Optimization-based Inference:} At inference time, the initial value of $\boldsymbol{z}$, denoted as $\boldsymbol{z}_0$, is sampled from the Gaussian distribution as Eq. \ref{random-init}. We keep the parameters of the decoder fixed and search for the optimal latent code $\boldsymbol{z}^*$ by gradient descent as Eq. \ref{neural_optim}. It is an iterative process as follows: 1) sampling an initial value $\boldsymbol{z}_0$; 2) decoding the corresponding robot motion; 3) calculating the loss of the objective function; 4) optimizing the latent code by gradient descent. In this way, we generate the best possible results for any unseen motion.

\vspace{-15pt}
\begin{gather}
    \boldsymbol{z}_0 \sim \mathcal{N}(0,\,\sigma^{2})\label{random-init} \\
    \boldsymbol{z}^*=\argmin_{\boldsymbol{z}}L(D,K(f_{\psi}(\boldsymbol{z})))+\frac{1}{\sigma^2}||\boldsymbol{z}||_2^2\label{neural_optim}
\end{gather}
\vspace{-10pt}

After searching for the optimal latent code for each motion, our method can provide promising results for unseen motions. However, it still has some shortcomings due to random initialization. It is time-consuming to start the optimization process from random values. Moreover, different ways of random initialization may lead to different convergence results, and it may easily fall into a bad local minimum without a good initial value.

\subsection{Neural Initialization} 


\noindent\textbf{Encoder As Initialization:} To overcome the shortcomings of random initialization, we introduce a parametrized encoder $f_{\phi}$ to provide a better initialization for faster and better convergence, where $\phi$ is the parameters of the encoder. The encoder takes the human demonstration as input and generates a better initial value $\boldsymbol{z}_0$, which should be close to the optimal latent code $\boldsymbol{z}^*$ so that latent optimization will take fewer iterations to converge and reduce the probability of falling into a bad local minimum. At training time, we optimize the parameters of the encoder \textbf{$f_{\phi}$} and the decoder $f_{\psi}$ simultaneously:

\begin{equation}
\label{training}
    \min_{\psi,\phi}L(D,K(f_{\psi}(f_{\phi}(D))))+\frac{1}{\sigma^2}||f_{\phi}(D)||_2^2
\end{equation}

At inference time, unlike feedforward networks that output the results directly, the encoder finds a good initial value $\boldsymbol{z}_0$ for each human demonstration as Eq. \ref{neural_init}. Then latent optimization starts from this initial value $\boldsymbol{z}_0$ and searches for the best latent code by gradient descent as Eq. \ref{neural_optim_new}. {The maximum number of iterations is set to 100, and the stopping criterion is that the maximum number of iterations is reached or the loss does not decrease in five consecutive iterations.}

\vspace{-15pt}
\begin{gather}\label{neural_init}
    \boldsymbol{z}_0=f_{\phi}(D)\\
    \boldsymbol{z}^*=\argmin_{\boldsymbol{z}}L(D,K(f_{\psi}(\boldsymbol{z})))+\frac{1}{\sigma^2}||\boldsymbol{z}||_2^2
    \label{neural_optim_new}
\end{gather}
\vspace{-15pt}



\subsection{Graph-based Encoder \& Decoder}

A common choice for designing the encoder and decoder is to concatenate the features of all joints and pass them through a fully connected network. However, the human and robot structures are rich in topological information, which may be lost in the fully connected layer. It is a more natural way to represent the human skeleton and robot structure as graphs, where each node corresponds to a joint of the human or robot, and each edge represents the connection relationship between the joints. The graph representation is invariant to node and edge permutations and has proved its effectiveness of extracting relevant information of the human body skeleton in many tasks \cite{yan2018spatial,li2020dynamic,zeng2021learning}. Its advantage is further discussed in the Appendix \cite{zhang2021kinematic}. In the task of motion retargeting, we choose to represent the data as graphs rather than vectors to better capture topological information.

Consider a graph $G=(V,E)$ that contains a set of $N$ nodes with feature dimension $C_1$, denoted by $V\in\mathbb{R}^{N\times{C_1}}$, and a set of $M$ edges with feature dimension $C_2$, denoted by $E\in\mathbb{R}^{M\times{C_2}}$. The graph may be a human skeleton graph or a robot structure graph. Let $\boldsymbol{x}_i$ be the feature of node $i$, $N(i)$ be the set of neighbor nodes of node $i$, and $\boldsymbol{e}_{j,i}$ be the edge feature from node $j$ to node $i$. 
For each neighbor node $j$ of node $i$, we first concatenate their node features and edge features, denoted as $\boldsymbol{z}_{i,j}=[\boldsymbol{x}_i,\boldsymbol{x}_j,\boldsymbol{e}_{j,i}]$. Then we obtain the message from each neighbor node by non-linear mapping and aggregate all the information by summation. Finally, we update the feature of node $i$ in a residual way by adding the aggregation result to the original feature. The update of node features is formulated as follows:

\begin{equation}
    {\boldsymbol{x}_{i}}^{'}=\boldsymbol{x}_i+\sum_{j\in{N(i)}}g(\boldsymbol{W}_{f}\boldsymbol{z}_{i,j}+\boldsymbol{b}_f)
\end{equation}

\noindent where $g$ is the LeakyReLU activation function, $\boldsymbol{W}_{f}$ and $\boldsymbol{b}_f$ are learnable parameters.

\begin{figure}[t]
\centering
\includegraphics[width=\linewidth]{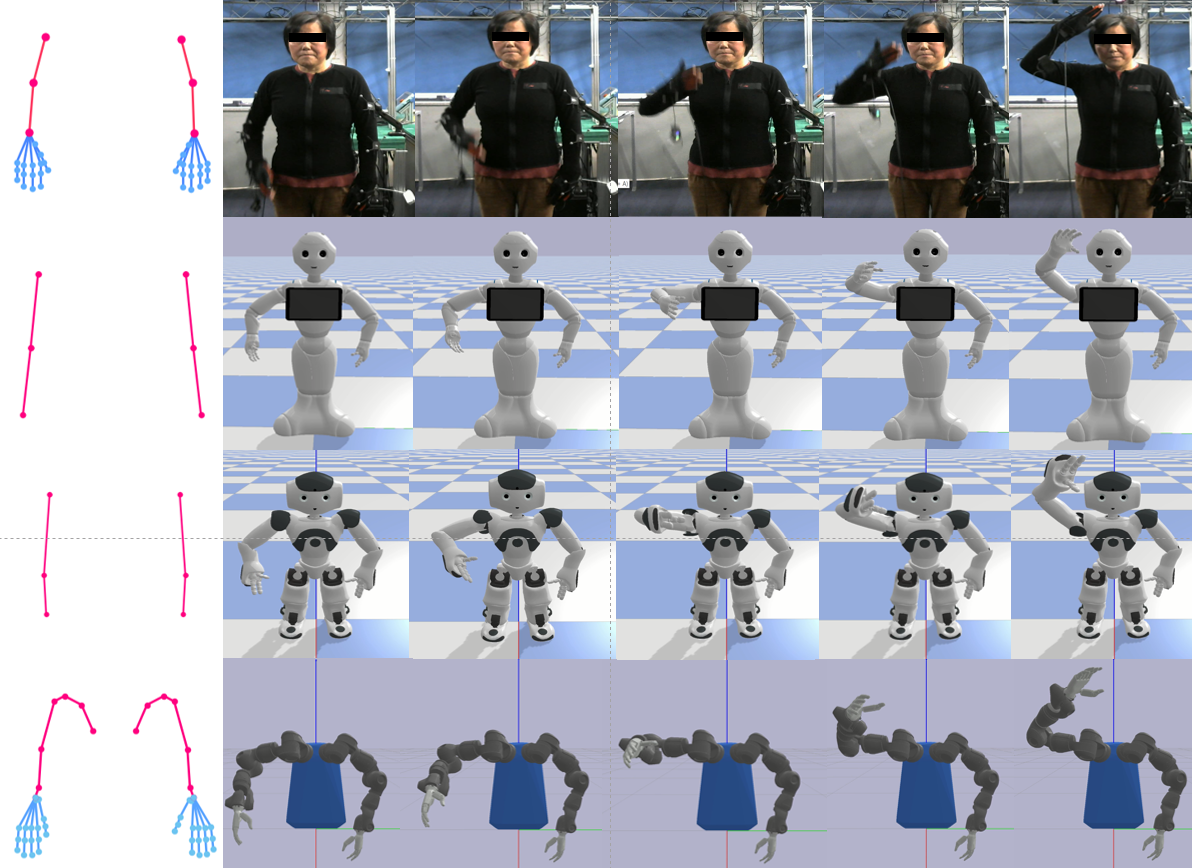}
\caption{Snapshots of “Can” sequence transferred by our proposed method on three different robots in the simulation environment. From the first row to the last row are the human demonstrator, Pepper, NAO, and YuMi, respectively. The first column is the corresponding human skeleton graph or robot structure graph.}
\label{sim-experiment}
\vspace{-5pt}
\end{figure}

The constructed human skeleton graph and robot structure graphs of three different robots are shown in Fig. \ref{sim-experiment}. Specifically, in the human skeleton graph, the input features of each node are the human joint positions and rotations, while the edge features are the offsets between human joints. On the other hand, in the robot structure graph, the output features of each node represent the rotation angle of the corresponding robot joint, while the edge features are the initial offsets and rotations between robot links. Both the human skeleton graph and the robot structure graph are directed graphs. We assume that there are two types of nodes, depending on whether the joint is located in the arm or in the hand. Nodes of the same type share the same graph convolution weights. By stacking multiple layers of graph convolution, the model can utilize the inherent information of the topological structure and learn deep-level features more effectively.

\begin{figure}[t]
\centering
\includegraphics[width=\linewidth]{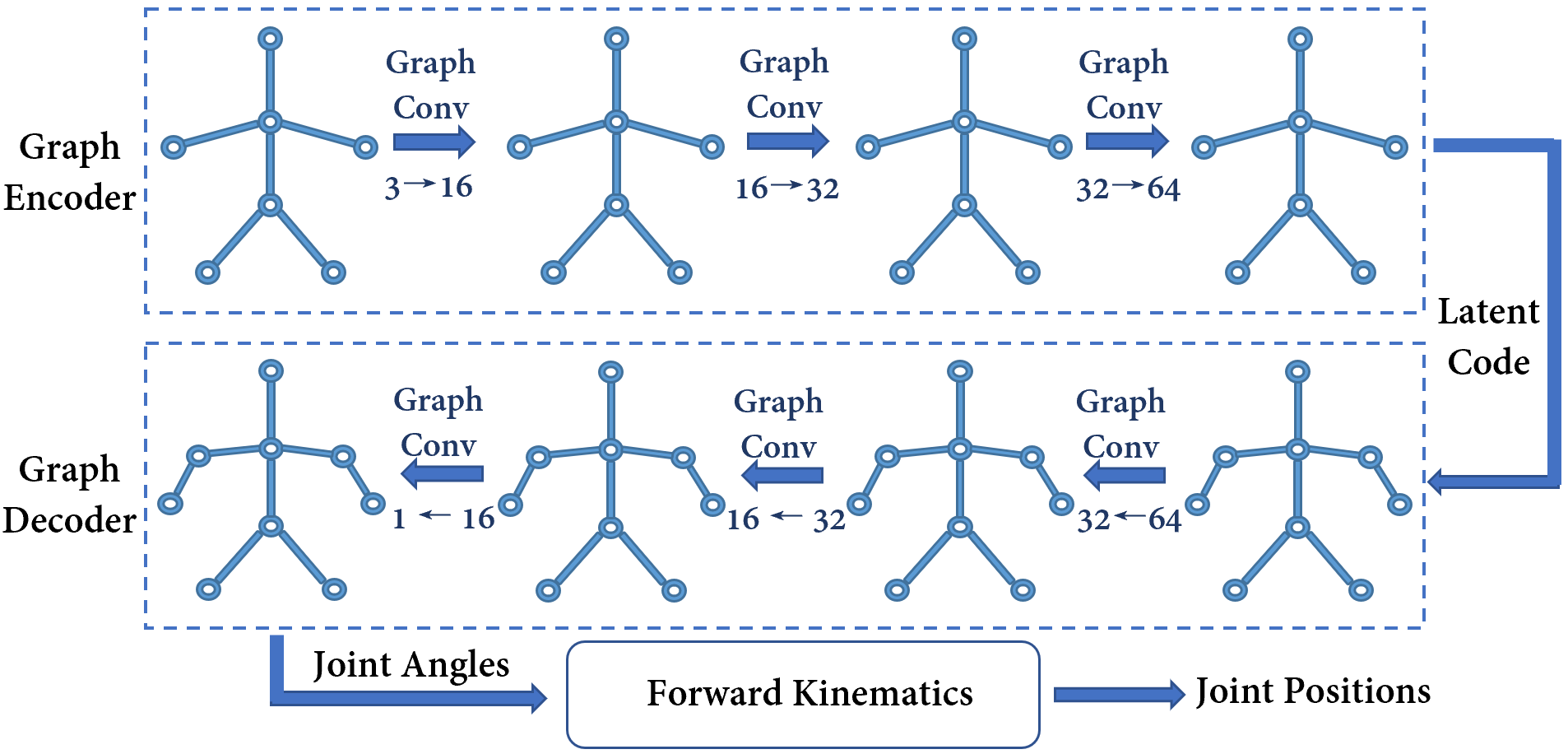}
\caption{Network architecture.}
\label{network}
\vspace{-18pt}
\end{figure}

{\subsection{Network Architecture}
The network architecture is shown in Fig. \ref{network}. The model consists of a graph encoder, a graph decoder, and a forward kinematics module. The graph encoder is composed of three layers of graph convolution, which extend the channels of the input features from 3 to 16, 32, and 64. Since the number of nodes in the human skeleton graph may differ from the robot structure graph, we use a linear layer to map the output features of the graph encoder to the latent code. The latent code is fed into the graph decoder after concatenation with the upper and lower limits of robot joints.}

The graph decoder is also composed of three layers of graph convolutions, which refine the channels of the output features to 32, 16, and 1. With the $tanh$ activation function, the output features of the decoder are constrained to the range -1 to 1. Then they are linearly remapped to the upper and lower limits of each robot joint. Thus each joint angle is ensured to be within the joint limits of the robot. Finally, the output joint angles are converted to joint positions using a forward kinematics layer. The forward kinematics is utilized as a differentiable but learning free network module, which allows us to optimize on the objective function. 



\section{Experiments}

\subsection{Experimental Setup}


We conduct experiments on motion retargeting from Chinese sign language to three different robots, including ABB's YuMi dual-arm robot, NAO, and Pepper. ABB's YuMi dual-arm robot contains 14 degrees of freedom and is equipped with Insipire-Robotics' dexterous hands, each with 6 degrees of freedom and 12 joints. For NAO and Pepper, we use only their robotic arms, which contain 10 degrees of freedom. The robot parameters are parsed from the unified robot description format, which includes the initial offsets, initial rotations, and upper and lower limits of the joint angles. The origin of the robot coordinate system is set at the center of its two shoulder joints, with the Z-axis in the vertical direction, the X-axis in the forward direction, and the Y-axis in the left-hand direction.

The Chinese sign language is performed by a professional sign language teacher whose shoulder, elbow, and wrist poses are captured by a high-precision optical system called OptiTrack. And the finger trajectories are obtained by data gloves called WiseGlove. The position data are normalized with a fixed coefficient and the rotation data are expressed in terms of Euler angles. The human coordinate system is set up in the same way as the robot. The total data contains sign language sequences of 5 daily scenarios with a total of 86 sequences and 13,847 frames. {The statistics of the collected dataset can be found in the Appendix \cite{zhang2021kinematic}.} We split the data into a training set and a test set, where the training set includes sequences from 3 scenarios, “Business”, “Railway Station” and “West Lake”, and the test set includes sequences from 2 scenarios, “Hospital” and “Introduction”. Therefore, the total training data contains 61 sign language sequences and 9,691 frames and the testing data contains 25 sign language sequences and 4,156 frames. The collected dataset will be open source soon. 

During training, we use a batch size of 16 and an Adam optimizer with a constant learning rate of 1e-4. The training process takes 1,301s until the model converges with an NVIDIA TITAN X GPU and an Intel(R) Xeon(R) E5-2696 CPU.

\subsection{Ablation Study}

\begin{figure}[t]
\centering
\includegraphics[width=0.9\linewidth]{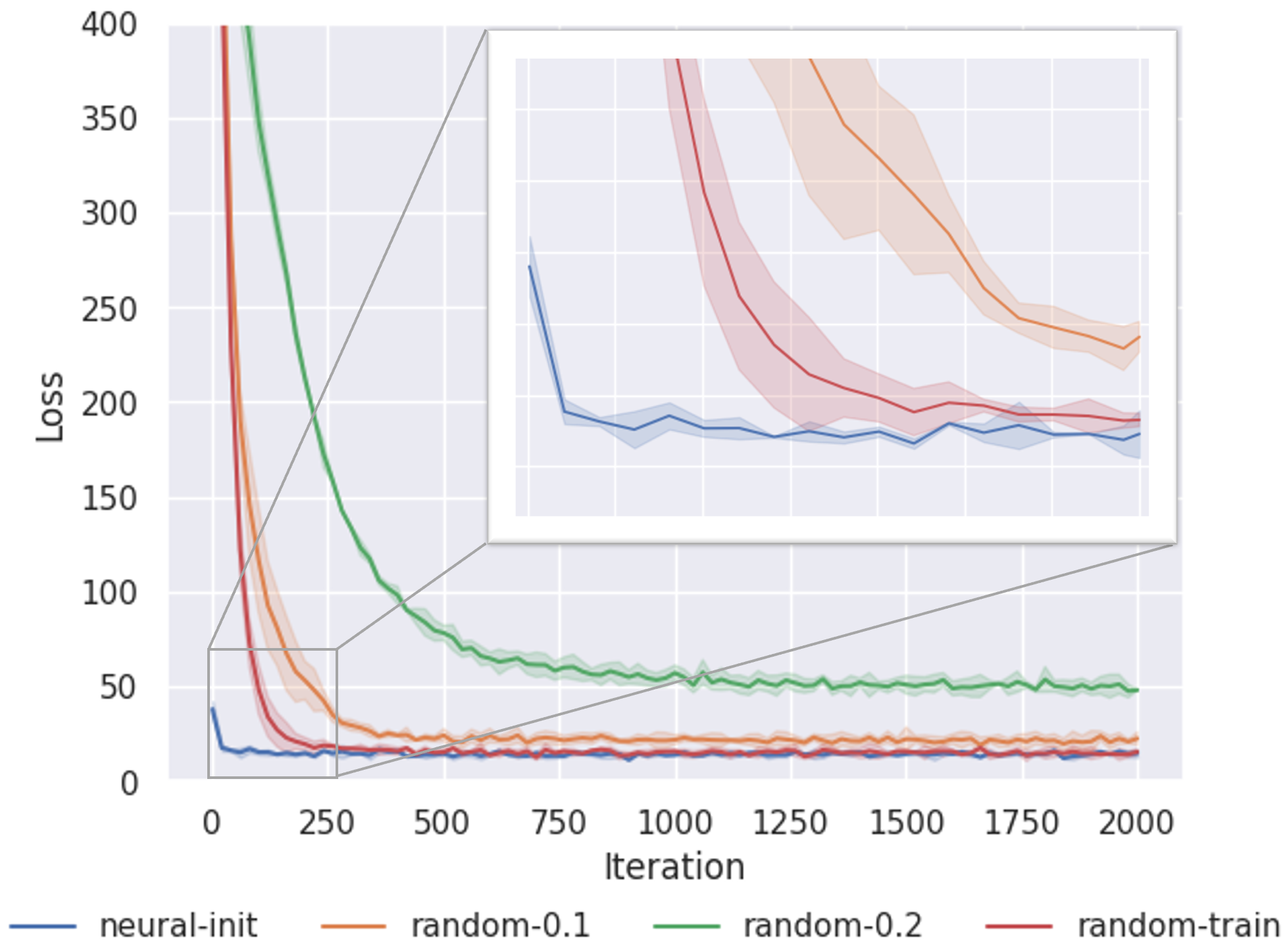}
\caption{Convergence results of neural initialization and random initialization of Gaussian distributions with a mean of 0 and a variance of 0.1, 0.2, and the training set's mean and variance.}
\label{training_curve}
\vspace{-15pt}
\end{figure}

\noindent\textbf{Initialization Selection:} To evaluate the benefit of using a deep graph encoder to provide initial values for latent vectors, we compare the performance of neural initialization and random initialization in the task of retargeting unseen motions in the test set to YuMi. We use three Gaussian distributions for random initialization as comparisons, two of which have a mean of 0 and a variance of 0.1, 0.2, and the other uses the mean and variance of the training set. From Fig. \ref{training_curve}, we could see that: 1) different initialization methods will lead to different convergence results; 2) random initialization may fall into a local minimum; 3) the deep graph encoder is able to provide a better initial value that helps to converge faster and better.

\begin{figure}[t]
\centering
\includegraphics[width=0.8\linewidth]{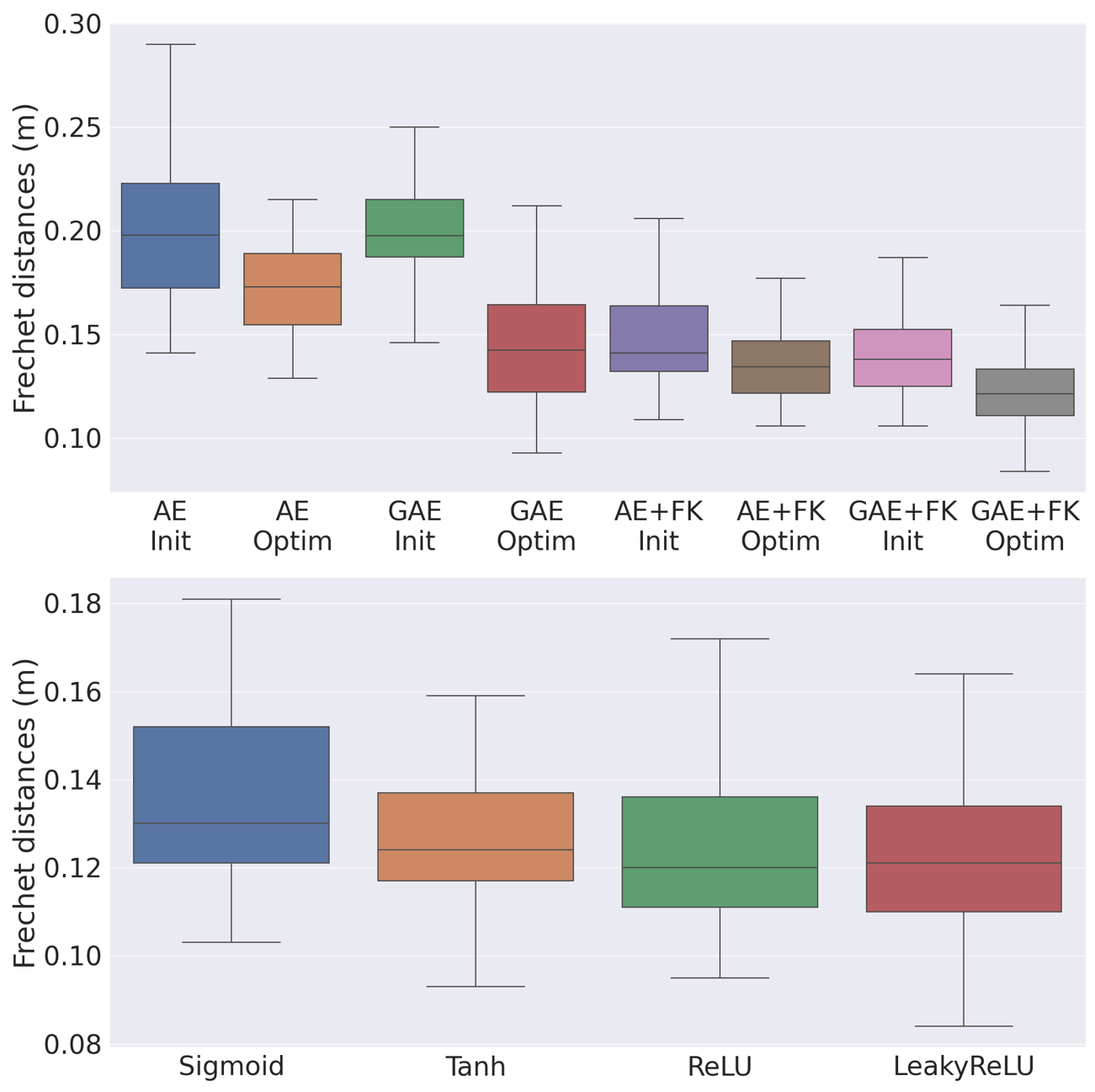}
\caption{{Frechet distances of different network architectures for unseen motions in the test set. More quantitative results can be found in the Appendix \cite{zhang2021kinematic}.}}
\label{ae_boxplot}
\vspace{-18pt}
\end{figure}


\noindent{\textbf{Network Architecture:} To validate the effectiveness of the network architecture, we conduct a comparative experiment on the use of graph representation, forward kinematics and latent optimization. We refer to the versions with and without graph representation as 'AE' and 'GAE'. 'FK' denotes the use of forward kinematics. 'Optim' and 'Init' indicates with and without latent optimization. Details can be found in the Appendix \cite{zhang2021kinematic}. We also compare the retargeting results of LeakyReLU with several common activation functions. The experiment is performed on the task of retargeting unseen motions in the test set to YuMi. From Fig. \ref{ae_boxplot}, we could see that: 1) the use of graph representation helps generalize better on unseen motions; 2) the use of forward kinematics as a differentiable but learning free network module helps optimize on the objective function; 3) optimization in the latent space further improves the performance; 4) Tanh, ReLU and LeakyReLU achieve similar results and LeakyReLU is better than the others.}

\begin{figure*}[htbp]
\centering
\includegraphics[width=0.8\linewidth]{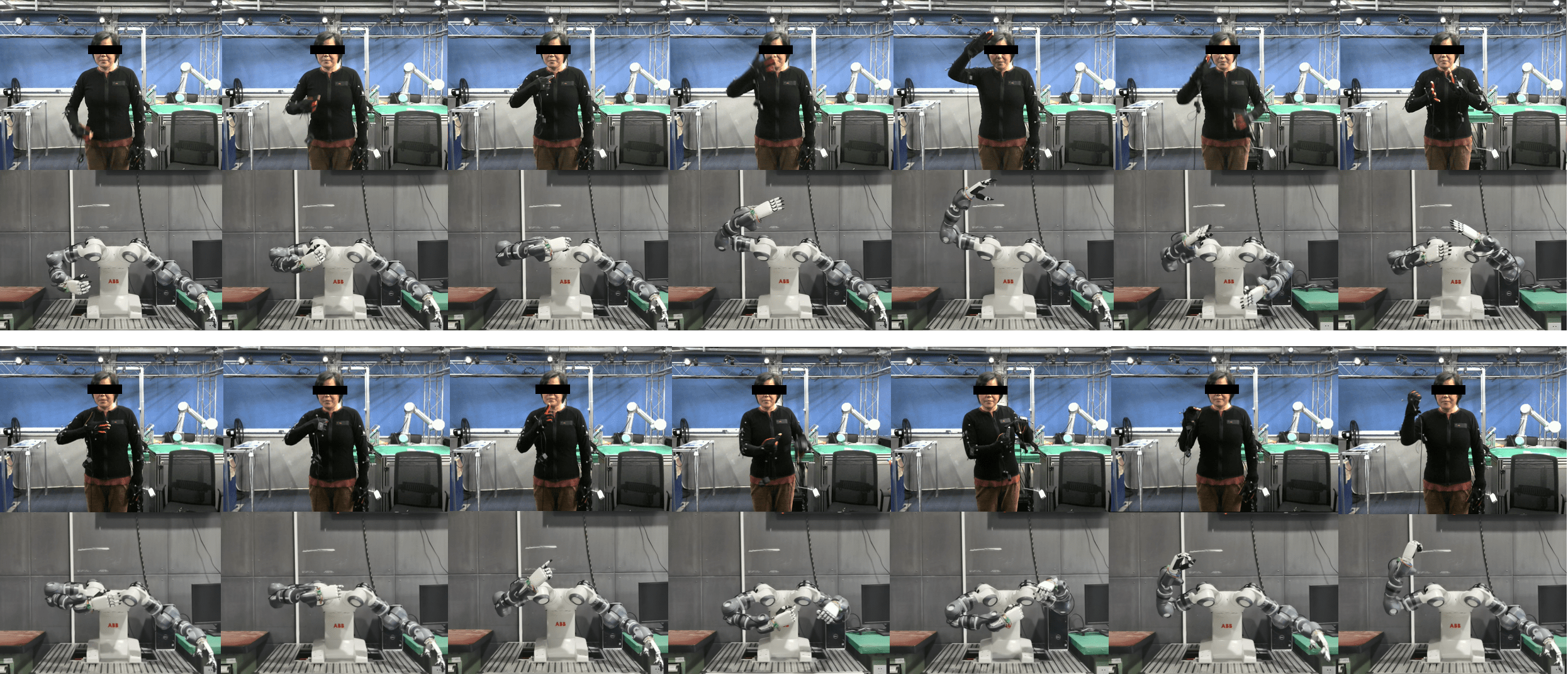}
\caption{Snapshots of motions transferred by our proposed method on the physical YuMi robot. The first two rows are “I can speak sign language” and the last two rows are “Give me this ability”.}
\label{real-experiment}
\vspace{-15pt}
\end{figure*}

\subsection{{Case Study}}


In Fig. \ref{sim-experiment}, we first demonstrate the results of motion retargeting from a human demonstrator to three different types of robots in the simulation environment built with PyBullet \cite{coumans2021}. Based on their structural proximity to the human demonstrator, the complexity of motion retargeting to these robots is ranked in ascending order as Pepper, Nao, and YuMi. Specifically, Pepper and Nao have fewer degrees of freedom and the proportion of their robot links is closer to the human, while YuMi is equipped with dexterous hands and has a more complex structure. The results show that our method can be applied to robots with different structures, which include different degrees of freedom, joint lengths, and joint limits. Furthermore, our method performs well not only on robots with relatively simple structures, but also on the robot with a significantly complex structure.


\subsection{{Comparative Study}}

To thoroughly evaluate the performance of our approach, we compare against four representative baselines on unseen motions in the test set:

\begin{itemize}

\item \textbf{DMPMR} \cite{liang2021dynamic} is an optimization-based method that combines graph optimization with Dynamic Movement Primitives (DMPs). It adopts a three-step optimization procedure to generate joint angles of robotic arms and calculates finger joint angles using linear mapping.

\item \textbf{NMG} \cite{choi2019towards} is an optimization-based pipeline that utilizes link length modifications and task space fine-tuning. It consists of four steps: descriptions preparation, motion optimization, inverse kinematics and post-processing.

\item \textbf{C-3PO} \cite{kim2020c} is a cyclic three-phase deep reinforcement learning method. It adopts the proximal policy optimization (PPO) \cite{schulman2017proximal} algorithm to learn the motion retargeting skill in quantitative and qualitative manners.

\item \textbf{NN} is a feedforward neural network that takes all joint poses of the human body as input and directly outputs the joint angles of the robot. It is trained in a supervised learning manner with the optimization results of DMPMR as ground truth, similar to \cite{choi2020nonparametric}.

\end{itemize}

\begin{figure}[t]
\centering
\includegraphics[width=0.9\linewidth]{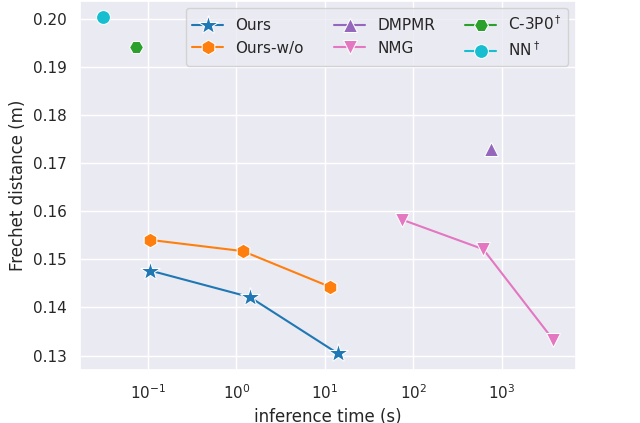}
\caption{Comparison between our neural latent optimization method and the baselines. We refer to our method as 'Ours' and the version without using the graph-based network design as 'Ours-w/o'. The closer the data points are to the lower left corner, the better the retargeting results, which means higher similarity is achieved with less inference time. Methods with the superscript $\dagger$ have produced infeasible robot motions in unseen motions of the test set. {More quantitative results can be found in the Appendix \cite{zhang2021kinematic}.}}
\label{comparison-figure}
\vspace{-12pt}
\end{figure}

We conduct an experiment to retarget unseen motions in the test set to the YuMi robot and compare the tracking error of the wrist and elbow as well as the optimization time. The tracking error is measured by the average Frechet distances \cite{eiter1994computing} of the wrist and elbow trajectories. The smaller the Frechet distance, the more similar the robot motion is to the demonstration. {Details of it can be found in the Appendix \cite{zhang2021kinematic}.} To better visualize the results, we compare both the average Frechet distance and the optimization time in Fig. \ref{comparison-figure}. We notice that our proposed method is able to provide promising results in unseen motions with less optimization time compared with optimization-based methods. Moreover, as the iterations of latent optimization increase, our method is able to achieve better motion retargeting results. We also observe that C-3PO and NN perform worse than optimization-based methods, since they may suffer from generalization to unseen motions, leading to some inaccurate or infeasible results. NN has the worst performance, which indicates that using the optimization output as the ground truth for training will lead to cumulative error consisting of optimization error and fitting error. 

%

\subsection{Real-World Qualitative Experiment}

We perform the real-world experiment on the ABB's YuMi dual-arm collaborative robot equipped with Insipire-Robotics' dexterous hands. The movement of the robotic arms and the dexterous hands is controlled by an external computer. As shown in Fig. \ref{real-experiment}, the robot motion generated by our proposed method is smooth and very close to that of the demonstrator. {The smoothness is further discussed in the Appendix \cite{zhang2021kinematic}.}

\section{CONCLUSIONS}

{This paper proposes a human-robot motion retargeting method that generates kinematically feasible robot motions based on neural latent optimization.} By optimizing in the latent space, our method can achieve better results compared to feedforward neural networks. With the better initialization provided by our deep graph encoder, our method can converge faster than previous optimization-based methods. Experimental results show that our method can generate smooth and similar motions that can be performed on physical robots. Future work will improve the limitation that it does not consider physical contact with the environment.

\vspace{-5pt}
\printbibliography

\newpage
\section{APPENDIX}

\subsection{Statistics of the Sign Language Dataset}

The statistics of the collected dataset are shown in Tab. \ref{statistics}. 

\subsection{Motivation of Using Latent Variables}

Due to the complex nonlinear objective function, optimization-based methods are prone to fall into a bad local minimum without a good initial value. The local minimum still exists when optimizing on the latent space, and we do not claim to solve this problem. However, we attempt to alleviate this problem by using latent variables. 

Firstly, the choice of initial values has a significant impact on the results of optimization and the speed of convergence. We introduce a graph encoder to provide a better initial value for optimization, which can reduce the probability of getting stuck in a bad local minimum and accelerate the optimization process. To evaluate the benefit of using a deep graph encoder to provide initial values for latent vectors, we compare the performance of neural initialization and random initialization in the task of retargeting unseen motions in the test set to YuMi. We use three Gaussian distributions for random initialization as comparisons, two of which have a mean of 0 and a variance of 0.1, 0.2, and the other use the mean and variance of the training set. From Fig. \ref{training_curve}, we could see that the deep graph encoder is able to provide a better initial value that helps to converge faster and better.

Secondly, solving an unconstrained optimization problem is simpler and more efficient than solving a constrained one. We are able to convert the constrained optimization problem into an unconstrained one by optimizing in the latent space instead of robot motion space. A nonlinear activation function is utilized to bound the output to a finite range. Then by linearly remap the output to the upper and lower limits of the robot's joint angles, we ensure that the joint limit constraints are satisfied. To thoroughly evaluate the performance of our approach, we compare against four representative baselines on unseen motions in the test set.
As the experimental results in Fig. \ref{comparison-figure} show, our method can achieve better motion retargeting results with both accuracy and efficiency, compared with other baselines.

Thirdly, similar ideas of optimizing in the latent space have also shown great performance in other areas, such as image generation and 3D geometry representation. \cite{park2019deepsdf} introduced to optimize the latent representation of 3D geometry. As far as we know, we are the first to explore the application of latent optimization in the task of motion retargeting and verify its feasibility, which would be beneficial to the community of motion retargeting.

\begin{table}[]
\centering
\caption{Statistics of the sign language dataset.}
\setlength{\tabcolsep}{4pt}
\setlength{\extrarowheight}{2pt}
\begin{tabular}{llcc}
\hline
Type & Scenario & \multicolumn{1}{l}{Number of Sequences} & \multicolumn{1}{l}{Number of Frames} \\ \hline
\multirow{3}{*}{Train} 
& Business        & 23 & 3671 \\
& Railway Station & 13 & 1756 \\
& West Lake       & 25 & 4264 \\ \hline
\multirow{2}{*}{Test}
& Hospital        & 12 & 2046 \\
& Introduction    & 13 & 2110 \\ \hline
\end{tabular}
\label{statistics}
\vspace{-15pt}
\end{table}

\subsection{Comparison with Kinematics}

The exact forward kinematics of the robot is also used in our method. Our model is composed of a graph encoder, a graph decoder, and forward kinematics as a differentiable but learning free network module. The graph decoder first maps the latent code $\boldsymbol{z}$ to the robot's joint angles. Then the forward kinematics module converts the joint angles to joint positions. Finally, we optimize the latent code by comparing the differences between the robot motion and human demonstration based on the objective function. The latent optimizatin process can be formulated as follows:

\begin{gather}
    \boldsymbol{z}^*=\argmin_{\boldsymbol{z}}L(D,K(f_{\psi}(\boldsymbol{z})))+\frac{1}{\sigma^2}||\boldsymbol{z}||_2^2
\end{gather}

\noindent where $\boldsymbol{z}$ be the latent code, $L$ be the objective function that measures the difference between robot motions and human demonstrations, $D$ be the poses of all human joints of the human demonstration, $K$ be the forward kinematics module, $f_{\psi}$ be the deep graph decoder and $\psi$ be the learnable parameters of the decoder.

We will explain the limitations of inverse kinematics and how our method avoids these limitations. Direct use of inverse kinematics of the robot is one way to address the problem of motion retargeting. However, there exists many limitations: 1) without considering the similarity of other joints, it only keeps the end effector positions of the robot consistent with those of the human; 2) it does not consider the kinematic constraints of the robot, such as joint limit constraints and collision avoidance; 3) it may produce multiple solutions when the robot has redundant degrees of freedom. \cite{liang2021dynamic} also shows that using inverse kinematics to perform motion retargeting will cause these problems and may lead to self-collisions. 

To solve these problems, a common approach is to consider the kinematic constraints and motion similarity as a secondary task and formulate it as an optimization problem. However, it will suffer from the same problems as other optimization-based methods \cite{liang2021dynamic,choi2019towards,wang2017generative}. It will take extensive time to optimize each motion and poor initialization may lead to a bad local minimum. As a comparison, our method exploits a graph decoder to establish the mapping relationship between the latent space and the robot motion space. The graph decoder ensures that the robot joint angles satisfy the joint limit constraints with the help of a nonlinear activation function. By optimizing the objective function in the latent space, it searches for the optimal solution that maximizes motion similarity and satisfies the kinematic constraints. To accelerate the optimization process and help converge to better results, a deep encoder is utilized to generate better initialization.

Compared with other networks, such as fully connected networks, the graph neural network can better capture human and robot topological information. Graphs are capable of modeling a set of objects and their relationship, such as social networks and physical systems \cite{zhou2020graph}. The human skeleton and robot structure can be naturally represented as graphs, which can model the properties of human or robot joints (nodes) and their connection relationships (edges). The commonly used vectorized representation treats all node information equally and neglects the rich relationship among nodes. For example, in the human skeleton, the left shoulder and left wrist have a more important influence on the left elbow, while other joints in the right arm do not. If we simply concatenate all joint information into a one-dimensional vector, the left shoulder and left wrist are considered the same as the other joints, and the rich connection relationship will be ignored. Therefore, we choose to represent the data as graphs rather than vectors. The ablation study of using graph neural network in Fig. \ref{comparison-figure} also verifies the effectiveness of the graph neural network.

\subsection{Comparison with the Classical Auto-encoder}

Our neural encoder and decoder have two major differences compared with the classical Auto-encoder and Variational Auto-encoder. One is that the forward kinematics is connected to the decoder as a differentiable but learning free network module, which allows us to optimize on the objective function. The other is that the encoder and decoder are constructed as graphs to better capture the inherent information of the topological structure.

First of all, for any unseen data, the proposed method utilizes forward kinematics as a differentiable but learning free network module to directly optimize on the objective function, which improves motion similarity and avoids self-collisions. However, the classical AE and VAE fit the optimization results during training as Eq. \ref{sup_loss} and inference directly by forward propagation as Eq. \ref{sup_inference}, which may perform poorly when encountering unseen motions.

\begin{gather}
    \label{sup_loss}
    L_{sup} = \sum_j{||r_j-r_j^*||_2^2} \\
    \label{sup_inference}
    r = f_{\omega}(D)
\end{gather}

\noindent where $L_{sup}$ be the supervised learning loss, $r_j$ be the output angle of robot joint $j$, $r_j^*$ be the ground truth provided by optimization-based methods, $D$ be the poses of all human joints of the human demonstration, $f_{\omega}$ be the network and $\omega$ be its learnable parameters.

The classical AE and VAE need a large amount of data for training to ensure that it generalizes well to unseen data. However, it is difficult to obtain large amounts of data, which takes lots of time and human effort. Since the quantity and diversity of training data is limited, the classical AE or VAE may not perform well in unseen motions, which results in inaccurate or infeasible robot motions. As a comparison, our method further optimizes the latent code for any unseen motion, which can improve the performance on unseen motions and combine the advantages of neural networks and optimization.

Secondly, the feature representation of the proposed method is different from the classical Auto-encoder or Variational Auto-encoder. They usually represent the inputs as one-dimensional vectors and use fully connected layers to learn the latent embedding of inputs. However, the human skeleton and robot structure can be naturally represented as graphs, which can model the properties of human or robot joints (nodes) and their connection relationships (edges). The use of vectorized feature representations may not be applicable on this task. Specifically, in the task of motion retargeting, the classical Auto-encoder or Variational Auto-encoder fails to capture the connection relationship of different joints in the human or robot structure, which is an important part of the topological information. As a comparison, our method models the human skeleton and robot structure as graphs, where each node corresponds to a joint of the human or robot, and each edge represents the connection relationship between the joints. In this way, we are able to capture the inherent information of the topological structure and have better generalization.

To thoroughly evaluate the performance of our approach, we compare against the classical Auto-encoder in the task of retargeting unseen motions in the test set to the YuMi robot.

\begin{itemize}

\item \textbf{AE} The classical Auto-encoder takes as input the concatenation of human joint positions and rotations and outputs joint angles of the robot. The encoder consists of three linear layers with 64, 128 and 256 channels, and the decoder consists of three linear layers with 128, 64 and 14 channels. It is trained in a supervised learning manner with the optimization results of DMPMR \cite{liang2021dynamic} as ground truth, similar to \cite{choi2020nonparametric}.

\item \textbf{GAE} The graph Auto-encoder uses the same model architecture as the graph encoder and decoder of our proposed method. The graph encoder is composed of three layers of graph convolution, which extend the channels of the input features from 3 to 16, 32, and 64. The graph decoder is also composed of three layers of graph convolutions, which refine the channels of the output features to 32, 16, and 1. It is trained in a supervised learning manner like AE.

\item \textbf{AE+FK} The Auto-encoder with forward kinematics uses the same model structure as the AE for the encoder and decoder, but with a different training approach. The forward kinematics is connected to the decoder as a differentiable but learning free network module, which allows it to optimize the objective function instead of fitting the optimization results.

\item \textbf{GAE+FK} The graph Auto-encoder with forward kinematics is the proposed method. The forward kinematics is connected to the decoder as a differentiable but learning free network module, which allows it to optimize the objective function instead of fitting the optimization results.

\end{itemize}

\begin{table}
\vspace{4pt}
\centering
\begin{threeparttable}[t]
\centering
\caption{Frechet distances of Auto-encoder, Graph Auto-encoder, Auto-encoder with forward kinematics, and Graph Auto-encoder with forward kinematics, denoted as AE, GAE, AE+FK, GAE+FK, respectively, for unseen motions in the test set. The results of each method include both with and without optimization, denoted as Optim and Init.}

\label{ae-table}
\setlength{\tabcolsep}{2pt}
\begin{tabular}{lccccccccc}
\hline
\multirow{3}{*}{Motion ID} & \multicolumn{8}{c}{Frechet Distance (m)} \\ \cline{2-9} 
& \multicolumn{2}{c}{AE} & \multicolumn{2}{c}{GAE} & \multicolumn{2}{c}{AE+FK} & \multicolumn{2}{c}{GAE+FK} \\ \cline{2-9} 
& Init      & Optim      & Init       & Optim      & Init        & Optim       & Init        & Optim        \\ \hline
1                    & 0.232 & 0.190 & 0.216 & 0.165 & 0.131 & 0.117 & 0.117 & \textbf{0.093} \\
2                    & $\:\;$0.222$^\P$ & 0.189 & 0.198 & 0.155 & 0.166 & 0.162 & 0.166 & \textbf{0.151} \\
3                    & 0.183 & 0.171 & 0.198 & 0.144 & 0.143 & 0.134 & 0.146 & \textbf{0.127} \\
4                    & 0.141 & 0.148 & 0.146 & 0.093 & 0.109 & 0.106 & 0.106 & \textbf{0.084} \\
5                    & 0.170 & 0.164 & 0.154 & 0.150 & 0.151 & 0.139 & 0.147 & \textbf{0.134} \\
6                    & 0.195 & 0.196 & 0.188 & 0.194 & 0.133 & 0.126 & 0.134 & \textbf{0.114} \\
7                    & 0.198 & 0.150 & $\:\;$0.212$^\P$ & 0.120 & 0.133 & \textbf{0.116} & 0.134 & 0.131 \\
8                    & 0.174 & 0.180 & 0.196 & 0.116 & 0.139 & 0.135 & 0.125 & \textbf{0.117} \\
9                    & 0.167 & 0.148 & 0.141 & 0.098 & 0.121 & 0.111 & 0.116 & \textbf{0.099} \\
10                   & 0.224 & 0.262 & 0.203 & 0.276 & 0.126 & 0.117 & 0.121 & \textbf{0.106} \\
11                   & 0.205 & 0.189 & 0.204 & 0.168 & 0.129 & 0.118 & 0.118 & \textbf{0.095} \\
12                   & 0.194 & 0.129 & 0.192 & 0.111 & 0.127 & 0.121 & 0.120 & \textbf{0.110} \\
13                   & 0.172 & 0.178 & 0.176 & 0.153 & 0.173 & 0.163 & 0.161 & \textbf{0.145} \\
14                   & 0.229 & 0.215 & 0.271 & 0.212 & 0.157 & 0.139 & 0.139 & \textbf{0.119} \\
15                   & 0.266 & 0.166 & 0.221 & 0.163 & 0.182 & 0.171 & 0.187 & \textbf{0.155} \\
16                   & 0.157 & 0.149 & 0.166 & 0.134 & 0.141 & 0.136 & 0.137 & \textbf{0.127} \\
17                   & 0.198 & 0.175 & 0.166 & 0.141 & 0.132 & 0.124 & 0.127 & \textbf{0.120} \\
18                   & 0.213 & 0.158 & 0.191 & \textbf{0.135} & 0.186 & 0.177 & 0.179 & 0.153 \\
19                   & 0.187 & 0.157 & 0.250 & \textbf{0.132} & 0.336 & 0.330 & 0.330 & 0.322 \\
20                   & 0.164 & 0.154 & 0.187 & 0.129 & 0.141 & 0.131 & 0.140 & \textbf{0.121} \\
21                   & 0.201 & 0.158 & 0.217 & \textbf{0.113} & 0.146 & 0.139 & 0.139 & 0.122 \\
22                   & 0.290 & 0.175 & 0.218 & \textbf{0.135} & 0.206 & 0.195 & 0.196 & 0.164 \\
23                   & 0.172 & 0.144 & 0.210 & 0.119 & 0.137 & 0.124 & 0.125 & \textbf{0.110} \\
24                   & 0.230 & 0.272 & 0.219 & 0.282 & 0.139 & 0.132 & 0.130 & \textbf{0.115} \\
25                   & 0.223 & 0.279 & 0.196 & 0.284 & 0.166 & 0.148 & 0.154 & \textbf{0.126} \\ \hline
Average              & 0.200 & 0.180 & 0.197 & 0.157 & 0.154 & 0.144 & 0.148 & \textbf{0.130} \\ \hline
\end{tabular}
\begin{tablenotes}
\item[\scalebox{1.0}{$\P$}] This superscript denotes infeasible robot motions with self-collision.
\end{tablenotes}
\end{threeparttable}
\vspace{-10pt}
\end{table}

We also compare the results of different models with and without latent optimization. The key points to influence the results of latent optimization are whether the initial values are good and whether the latent space captures the distribution of unseen motions. The optimization result depends mainly on the cost of the objective function. From Tab. \ref{ae-table}, we could see that: 1) the classical AE performs poorly on unseen data in the test set, which suggests that using the optimization results as ground truth for training will lead to cumulative error consisting of optimization error and fitting error; 2) the use of forward kinematics as a differentiable but learning free network module helps improve the performance of motion retargeting; 3) the use of graph representation helps generalize better on unseen motions; 4) optimization in the latent space further improves the performance of motion retargeting. The experimental results validate the advantages of our proposed method, which utilizes graph representations and combines with a forward kinematics layer to perform further optimization.

\subsection{Advantage of Graph Representation}
Different network architectures have different relational inductive biases. The inductive biases express assumptions about the space of solutions. Common deep learning building blocks include fully connected layer, convolutional layers, and graph networks. The fully connected layer has no sharing in weights and assumes that the relationship of input units is all-to-all. The inductive bias of the fully connected layer is very weak, so more data are needed for training. With the same amount of data, the fully connected layer may generalize relatively poorly, which also explains the results in the previous table. The convolutional layer reuses a local kernel function multiple times across the input and is invariant to the spatial translation. It assumes the locality of grid elements, while the human skeleton data are difficult to represent in grid form. The graph network considers the joints as nodes and their connections as edges. It is invariant to node and edge permutations. The relational inductive bias is more consistent with the structure of humans and robots.

The human skeleton and robot structure can be naturally represented as graphs, which can model the properties of human or robot joints (nodes) and their connection relationships (edges). The commonly used vectorized representation treats all node information equally and neglects the rich relationship among nodes. For example, in the human skeleton, the left shoulder and left wrist have a more important influence on the left elbow, while other joints in the right arm do not. If we simply concatenate all joint information into a one-dimensional vector, the left shoulder and left wrist are considered the same as the other joints, and the rich connection relationship will be ignored. Since the human skeleton and robot structure is closer to the inductive bias of graph networks, we can generalize better to unseen motions with the same amount of data. Therefore, we choose to represent the data as graphs rather than vectors to to better capture human and robot topological information.

From experimental results in Tab. \ref{ae-table}, we also could see that the use of graph representation helps generalize better to unseen motions. The graph neural network has also been successfully applied in other skeleton-based tasks, such as action recognition and motion prediction. \cite{yan2018spatial} proposed a spatial-temporal graph convolutional network to learn the significant information of human body skeletons for action recognition. \cite{li2020dynamic} introduced a dynamic multiscale graph neural network to predict 3D skeleton-based human motions. \cite{zeng2021learning} presented a skeletal graph neural network that is robust and effective for 3D pose estimation. These works also illustrate the advantages of using graph neural networks in extracting relevant information about human motions.

\subsection{Metric Definition}

The tracking error in our experiments is measured by the average Frechet distances of the wrist and elbow trajectories, because the shoulder position of human skeleton is almost fixed and the elbow and wrist contribute more to the motion similarity. We first calculate the Frechet distance at the wrist and elbow separately, and then take their average as the final tracking error.

The Frechet distance is first proposed to measure the similarity between two trajectories \cite{eiter1994computing}. It measures the maximum distance between two trajectories and its mathematical form is shown in Eq. \ref{frechet}. Let A and B be two given trajectories, the Frechet distance is defined as the infimum over all reparameterizations $\alpha$ and $\beta$ of [0, 1] of the maximum over all $t\in[0,1]$ of the distance between $A(\alpha(t))$ and $B(\beta(t))$:

\begin{equation}
\label{frechet}
    F(A,B) = \inf_{\alpha,\beta}\max_{t\in[0,1]}\{d(A(\alpha(t)),B(\beta(t)))\}
\end{equation}

The smaller the Frechet distance, the more similar the trajectories are. It has also been applied in the baseline DMPMR \cite{liang2021dynamic} to measure the similarity of human demonstrations and robot motions in the task of sign language motion retargeting.

\begin{figure*}[htbp]
\centering
\includegraphics[width=0.9\linewidth]{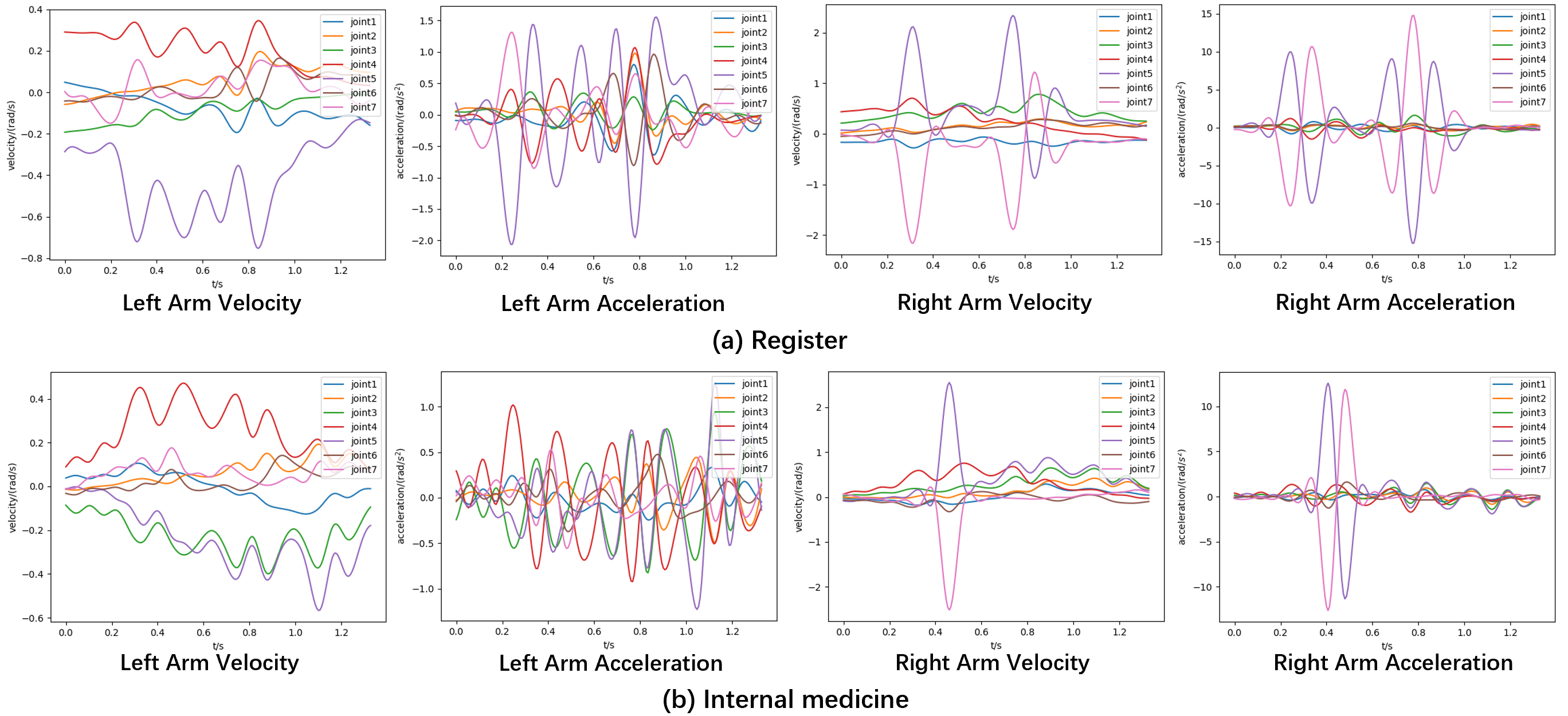}
\caption{Velocity and acceleration trajectories of robot joints. We have taken a part of the whole trajectory that is rich in variations.}
\label{vel-acc-traj-6}
\end{figure*}

\subsection{Smoothness Evaluation}

We just observe that the robot motion generated by the proposed method is smooth and close to that of the demonstrator in the real-world experiment. There is no mechanism in our method to ensure smoothness. We explain our smoothness by the similarity between the generated motion and the inherently smooth human demonstrations.

\begin{figure*}[htbp]
\centering
\includegraphics[width=0.9\linewidth]{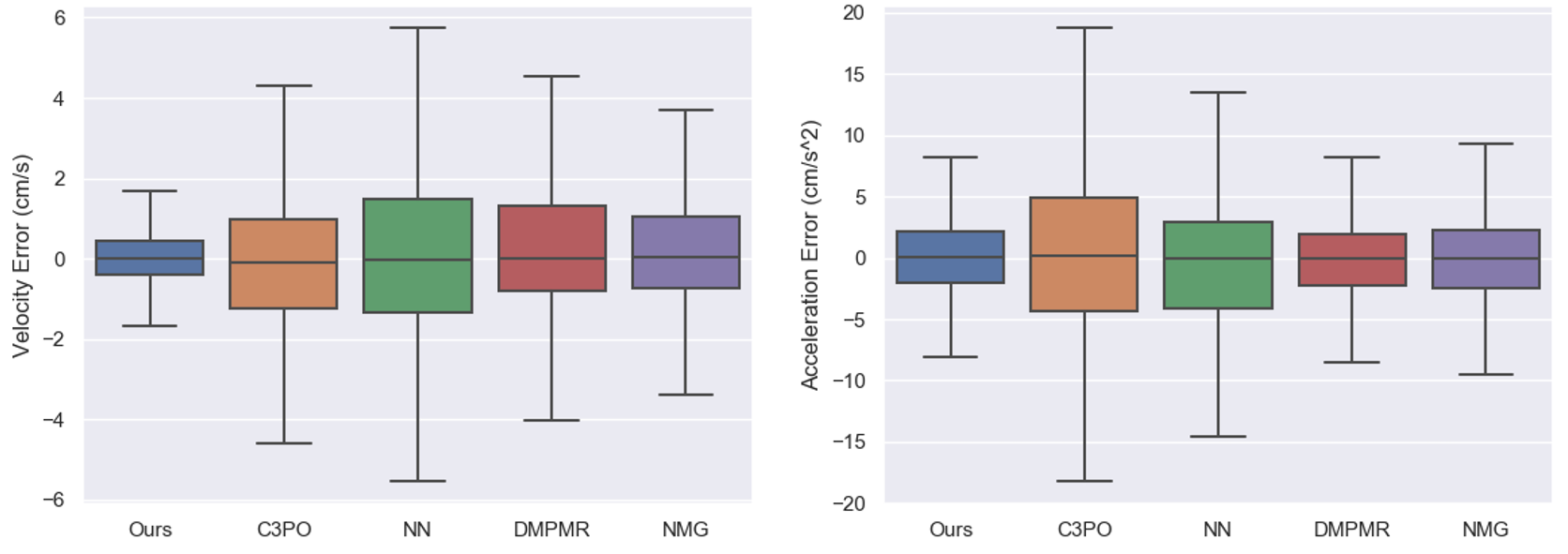}
\caption{Velocity and acceleration error of different methods.}
\label{vel-error-7}
\vspace{-15pt}
\end{figure*}

To provide a more thorough evaluation of the smoothness, we visualize the velocity and acceleration trajectories of all robot joints for some sign language motions in Fig. \ref{vel-acc-traj-6}. We further compare the velocity error of our method with the baselines. The velocity error is calculated by the difference in velocity between the human and robot wrists and elbows. From Fig. \ref{vel-error-7}, we could see that the proposed method has the smallest velocity error compared with other baselines, which indicates that the robot motions generated by our method are close to the human demonstrator. Since the human movements are inherently smooth, we believe that the similarity to human velocity accounts for the smoothness of the generated robot motions. 

We also compare the acceleration error of our method with the baselines in Fig. \ref{vel-error-7}. The acceleration error is calculated by the difference in acceleration between the human and robot wrists and elbows. DMPMR has the smallest acceleration error because it explicitly takes into account the smoothness. The smoothness is not explicitly considered in the objective function of our method and NMG, nor the reward design of C3PO. However, our acceleration error is smaller, which we believe is because our method generates robot motions similar to the inherently smooth human movements.

\begin{figure*}[htbp]
\centering
\includegraphics[width=0.8\linewidth]{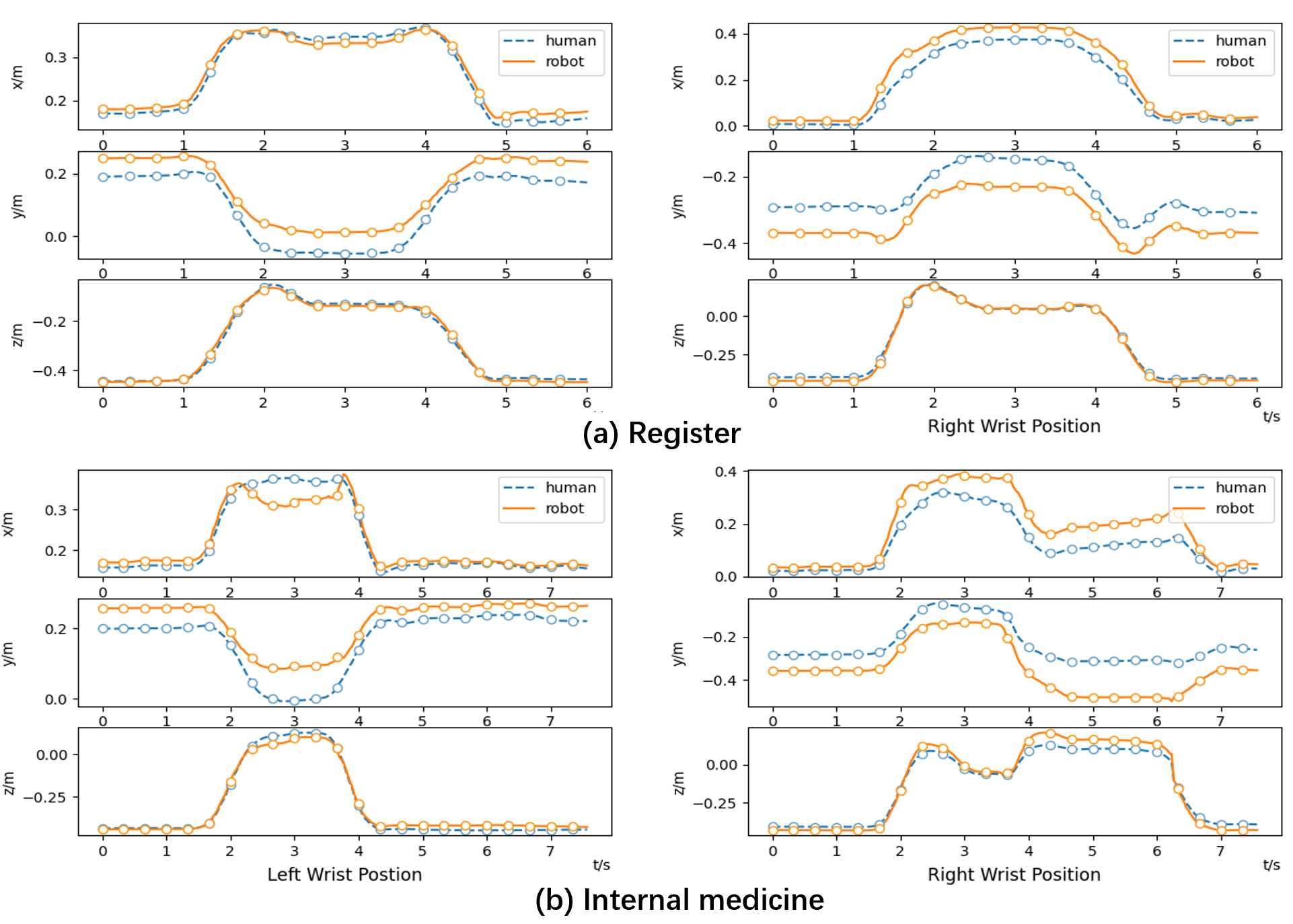}
\caption{Position trajectories of the human and robot wrists.}
\label{trajectory-7}
\vspace{-10pt}
\end{figure*}

We visualize the joint position trajectories of the human and robot wrists in Fig. \ref{trajectory-7}. We demonstrate the retargeting results of two different sign language motions from the human demonstrator to the YuMi robot. The difference between robot and human movements may be caused by dual-arm collision avoidance. The trajectories show that the robot movements we generate are close to the human demonstrator.

Thanks for the reviewer's valuable suggestion. In future work, we will further extend our approach to the temporal domain to explicitly take into account the smoothness.

\subsection{Comparison with Other Motion Retargeting Methods}

We believe that task-based motion retargeting could be categorized into kinematics-based \cite{choi2020nonparametric,kim2020c,choi2019towards} and dynamics-based \cite{peng2018deepmimic,peng2018sfv,ayusawa2017motion} motion retargeting. We focus on the task of sign language motion retargeting, which belongs to the kinematics-based one. The dynamics of the robot is not taken into account by our approach. The objective function only considers motion similarity and kinematic constraints. Different from other tasks, such as humanoid walking, which has rich physical contact with the environment, sign language motion retargeting completely focuses on the consistency of the kinematics of the robot. The task is defined as the similarity of the joints, thus less attention has been paid to the dynamics. To maintain the visual similarity of the robot motion, we scale the joint positions just as other motion retargeting methods \cite{choi2019towards,choi2020nonparametric} have done, to take into account the differences in movement trajectories due to inconsistencies in robot and human structures. If the method is to be used for other tasks, such as object grasping, there is no need to introduce normalization. Further consideration of task objectives may need to modify the objective functions to adapt to specific tasks.

\begin{figure}[ht]
\centering
\includegraphics[width=0.95\linewidth]{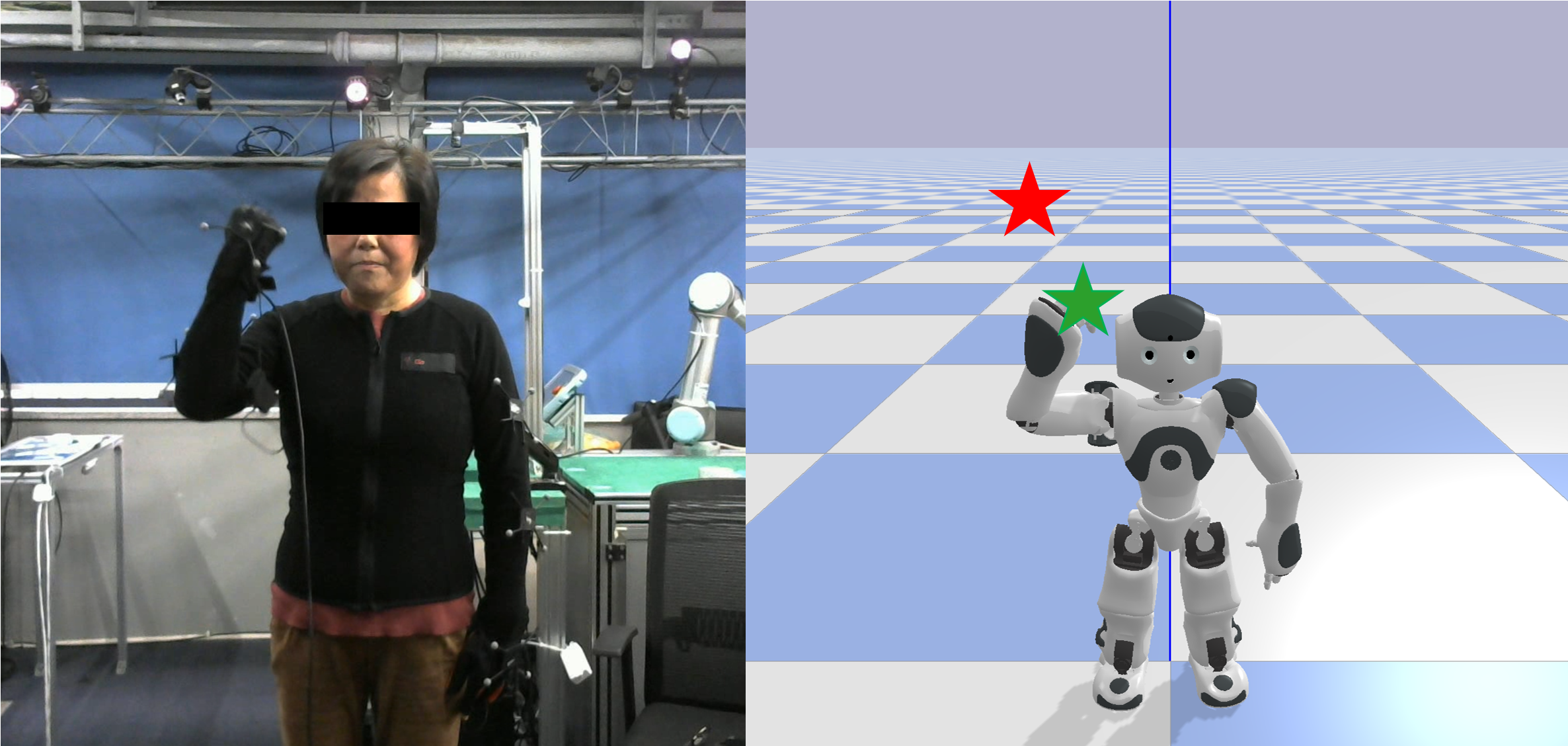}
\caption{End effector position with and without normalization. On the left is the human demonstration, and on the right is the robot motion. The green and red stars indicate the end effector positions with and without normalization, respectively.}
\label{normalization}
\end{figure}

To better illustrate the important role of normalization in the task of sign language motion retargeting, we take the example of the NAO robot, which is much smaller than the size of the human. If we use the absolute position of the human end effector as a reference, it may exceed the robot workspace. As shown in Fig. \ref{normalization}, the reference position without normalization, denoted by the red star, may lead to distortion of the movements, while the reference position with normalization, denoted by the green star, scales according to the size of the robot and can better maintain the similarity of the movements.

\subsection{Generalization to Other Demonstrators or Robots}

The model used in this experiment is not the same for all robots. There are several different cases here, and we will discuss each case separately as follows:

\textbf{Case 1:} The number of robot joints varies. We need to train a separate model for each robot, because each robot has different degrees of freedom, resulting in different output dimensions. For example, NAO has 10 degrees of freedom with an output dimension of 10, while YuMi has 14 degrees of freedom with an output dimension of 14. They need to train a separate model.

\textbf{Case 2:} Robots have the same number of joints but different joint lengths. Then the network edge feature values will change and need to be retrained. If we want the generalization of joint length, we need to have multiple robots with different lengths but the same number of joints in the training set, thus providing enough diversity in the edge features to facilitate generalization. Since the current network is one demonstrator to one robot, we cannot generalize directly.

\textbf{Case 3:} The demonstrators are different. Then the edge feature value will change and the network needs to be retrained. However, it is possible to apply the trained model to other demonstrators, since different demonstrators have the same joint number and input dimension. If we want to help the model generalize better over different demonstrators, we need to enrich the number of demonstrators in the dataset.

\begin{table}[]
\begin{threeparttable}[t]
\centering
\caption{Frechet distances of the proposed method and the baselines for unseen motions in the test set. The Frechet distance measures the similarity between the robot motion and the human demonstration. A smaller Frechet distance indicates a more similar robot motion.}
\label{comparison-table}
\setlength{\tabcolsep}{4.5pt}
\begin{tabular}{lccccc}
\hline
\multicolumn{1}{c}{\multirow{2}{*}{Motion}} & \multicolumn{5}{c}{Frechet Distance (m)} \\ \cline{2-6} 
\multicolumn{1}{c}{}                      & DMPMR  & NMG    & C-3PO & NN    & Ours  \\ \hline
Can                     & 0.121 & 0.127 & 0.187 & 0.232 & \textbf{0.093} \\
University              & 0.328 & \textbf{0.148} & $\:\;$0.197$^\P$ & $\:\;$0.222$^\P$ & 0.151 \\
Lab                     & 0.158 & 0.154 & $\:\;$0.203$^\P$ & 0.183 & \textbf{0.127} \\
I                       & 0.109 & 0.119 & 0.180 & 0.141 & \textbf{0.084} \\
Speak sign language     & 0.147 & 0.143 & $\:\;$0.216$^\P$ & 0.170 & \textbf{0.134} \\
Am                      & \textbf{0.092} & 0.131 & 0.181 & 0.195 & 0.114 \\
Robot                   & 0.341 & 0.136 & $\:\;$0.198$^\P$ & 0.198 & \textbf{0.131} \\
Zhejiang                & 0.180 & 0.128 & $\:\;$0.209$^\P$ & 0.174 & \textbf{0.117} \\
YuMi                    & 0.132 & 0.142 & 0.190 & 0.167 & \textbf{0.099} \\
Of                      & 0.123 & 0.144 & 0.198 & 0.224 & \textbf{0.106} \\
Give                    & 0.128 & 0.116 & 0.200 & 0.205 & \textbf{0.095} \\
Ability                 & 0.117 & 0.133 & 0.196 & 0.194 & \textbf{0.110} \\
This                    & \textbf{0.137} & 0.138 & $\:\;$0.209$^\P$ & 0.172 & 0.145 \\
Uncomfortable           & 0.128 & 0.127 & 0.198 & 0.229 & \textbf{0.119} \\
Internal medicine       & 0.366 & \textbf{0.136} & $\:\;$0.193$^\P$ & 0.266 & 0.155 \\
Where                   & 0.138 & 0.130 & 0.176 & 0.157 & \textbf{0.127} \\
Like                    & 0.144 & 0.139 & 0.179 & 0.198 & \textbf{0.120} \\
Help                    & 0.147 & \textbf{0.133} & $\:\;$0.188$^\P$ & 0.213 & 0.153 \\
Volunteer               & \textbf{0.126} & 0.138 & $\:\;$0.186$^\P$ & 0.187 & 0.322 \\
Find                    & 0.140 & 0.128 & 0.186 & 0.164 & \textbf{0.121} \\
Register                & 0.251 & 0.127 & 0.225 & 0.201 & \textbf{0.122} \\
Queue                   & 0.135 & \textbf{0.123} & 0.200 & 0.290 & 0.164 \\
Have a problem          & 0.144 & 0.127 & 0.184 & 0.172 & \textbf{0.110} \\
Window                  & 0.151 & 0.133 & 0.185 & 0.230 & \textbf{0.115} \\
Self-service machine    & 0.337 & 0.130 & $\:\;$0.190$^\P$ & 0.223 & \textbf{0.126} \\ \hline
Average                 & 0.173 & 0.133 & 0.194 & 0.200 & \textbf{0.130} \\ \hline
\end{tabular}
\begin{tablenotes}
\item[\scalebox{1.0}{$\P$}] This superscript denotes infeasible robot motions with self-collision.
\end{tablenotes}
\end{threeparttable}
\vspace{-15pt}
\end{table}

\begin{table}
\centering
\begin{threeparttable}[t]
\centering
\caption{Frechet distances of different activation functions for unseen motions in the test set. A smaller Frechet distance indicates a more similar robot motion.}
\label{activation-table}
\setlength{\tabcolsep}{6pt}
\begin{tabular}{lccccc}
\hline
\multicolumn{1}{c}{\multirow{2}{*}{Motion}} & \multicolumn{4}{c}{Frechet Distance (m)} \\ \cline{2-5} 
\multicolumn{1}{c}{}    & Sigmoid & Tanh & ReLU & LeakyReLU  \\ \hline
Can                     & 0.113 & 0.110 & 0.098 & \textbf{0.093} \\
University              & 0.156 & 0.153 & 0.151 & \textbf{0.151} \\
Lab                     & 0.130 & 0.126 & \textbf{0.125} & 0.127 \\
I                       & 0.103 & 0.093 & 0.095 & \textbf{0.084} \\
Speak sign language     & 0.152 & 0.137 & 0.136 & \textbf{0.134} \\
Am                      & 0.121 & 0.120 & 0.116 & \textbf{0.114} \\
Robot                   & 0.128 & 0.120 & \textbf{0.116} & 0.131 \\
Zhejiang                & 0.131 & 0.128 & 0.127 & \textbf{0.117} \\
YuMi                    & 0.115 & 0.105 & 0.104 & \textbf{0.099} \\
Of                      & 0.121 & 0.110 & 0.110 & \textbf{0.106} \\
Give                    & 0.120 & 0.105 & 0.106 & \textbf{0.095} \\
Ability                 & 0.125 & 0.117 & 0.110 & \textbf{0.110} \\
This                    & 0.157 & 0.146 & 0.152 & \textbf{0.145} \\
Uncomfortable           & 0.138 & 0.124 & 0.120 & \textbf{0.119} \\
Internal medicine       & 0.161 & 0.168 & 0.172 & \textbf{0.155} \\
Where                   & 0.132 & 0.133 & 0.128 & \textbf{0.127} \\
Like                    & 0.132 & 0.121 & \textbf{0.119} & 0.120 \\
Help                    & 0.169 & 0.148 & 0.154 & \textbf{0.153} \\
Volunteer               & 0.362 & 0.317 & \textbf{0.302} & 0.322 \\
Find                    & 0.130 & 0.132 & 0.128 & \textbf{0.121} \\
Register                & 0.123 & 0.120 & \textbf{0.116} & 0.122 \\
Queue                   & 0.181 & \textbf{0.159} & 0.168 & 0.164 \\
Have a problem          & 0.121 & 0.115 & 0.111 & \textbf{0.110} \\
Window                  & 0.128 & 0.122 & 0.117 & \textbf{0.115} \\
Self-service machine    & 0.145 & 0.125 & 0.129 & \textbf{0.126} \\ \hline
Average                 & 0.144 & 0.134 & 0.132 & \textbf{0.130} \\ \hline
\end{tabular}
\end{threeparttable}
\vspace{0pt}
\end{table}

\end{document}